\documentclass{article} 
\usepackage{iclr2017_conference,times}
\usepackage[
colorlinks,
linkcolor = blue,
citecolor=blue,
urlcolor=blue,
filecolor=black,
pdfauthor={Casper Kaae Sonderby, Jose Caballero, Lucas Theis, Wenzhe Shi and Ferenc Huszar},
pdftitle={Amortised MAP Inference for Image Super-Resolution},
pdfdisplaydoctitle=true]{hyperref}
\usepackage{url}
\usepackage[export]{adjustbox}
\usepackage{subcaption}
\usepackage{amsmath,amsfonts,amsthm,amssymb} 
\usepackage{wrapfig}
\usepackage[normalem]{ulem}
\usepackage{natbib}
\usepackage{graphicx} 
\usepackage[textsize=small]{todonotes}
\usepackage[T1]{fontenc}
\usepackage[font=small,labelfont=bf]{caption}
\usepackage[utf8]{inputenc}
\usepackage{placeins}
\usepackage{tikz}
\usepackage{pgfplots}
\usepackage{enumitem}
\usepackage{titlesec}

\iclrfinalcopy

\titlespacing*{\subsection}
{0pt}{2ex}{0.0ex}

\setlist[itemize]{noitemsep, nolistsep, leftmargin=*}
\setlist[enumerate]{noitemsep, nolistsep, leftmargin=*}

\DeclareCaptionLabelFormat{andtable}{#1~#2  \&  \tablename~\thetable}
\DeclareMathOperator*{\argmin}{\arg\!\min}
\DeclareMathOperator*{\argmax}{\arg\!\max}

\newcommand{\affgan}[0]{AffGAN}
\newcommand{\affdenoiser}[0]{AffDG}
\newcommand{\affdensity}[0]{AffLL}
\newcommand{\affmse}[0]{AffMSE}

\newcommand{\softdenoiser}[0]{SoftDG}
\newcommand{\softgan}[0]{SoftGAN}

\title{Amortised MAP Inference for\\ Image Super-resolution}

\author{Casper Kaae S\o nderby\textsuperscript{1\,2}\thanks{Work done while CKS was an intern at Twitter}\,, Jose Caballero\textsuperscript{1}\!, Lucas Theis\textsuperscript{1}\!, Wenzhe Shi\textsuperscript{1}\& Ferenc Husz\'{a}r\textsuperscript{1}\\
\texttt{casperkaae@gmail.com, \{jcaballero,ltheis,wshi,fhuszar\}@twitter.com} \\
\textsuperscript{1}Twitter, London, UK\\
\textsuperscript{2}University of Copenhagen, Denmark\\
}

\begin{document}

\maketitle
\begin{abstract}
Image super-resolution (SR) is an underdetermined inverse problem, where a large number of plausible high resolution images can explain the same downsampled image. Most current single image SR methods use empirical risk minimisation, often with a pixel-wise mean squared error (MSE) loss.
However, the outputs from such methods tend to be blurry, over-smoothed and generally appear implausible.
A more desirable approach would employ Maximum a Posteriori (MAP) inference, preferring solutions that always have a high probability under the image prior, and thus appear more plausible. Direct MAP estimation for SR is non-trivial, as it requires us to build a model for the image prior from samples. Here we introduce new methods for \emph{amortised MAP inference} whereby we calculate the MAP estimate directly using a convolutional neural network. We first introduce a novel neural network architecture that performs a projection to the affine subspace of valid SR solutions ensuring that the high resolution output of the network is always consistent with the low resolution input. Using this architecture, the amortised MAP inference problem reduces to minimising the cross-entropy between two distributions, similar to training generative models. We propose three methods to solve this optimisation problem: (1) Generative Adversarial Networks (GAN) (2) denoiser-guided SR which backpropagates gradient-estimates from denoising to train the network, and (3) a baseline method using a maximum-likelihood-trained image prior. Our experiments show that the GAN based approach performs best on real image data. Lastly, we establish a connection between GANs and amortised variational inference as in e.\,g.\ variational autoencoders.
\end{abstract}

\section{Introduction\label{sec:introduction}}

Image super-resolution (SR) is the underdetermined inverse problem of estimating a high resolution (HR) image given the corresponding low resolution (LR) input. This problem has recently attracted significant research interest due to the potential of enhancing the visual experience in many applications while limiting the amount of raw pixel data that needs to be stored or transmitted. While SR has many applications in for example medical diagnostics or forensics \citep[][and references therein]{nasrollahi2014super}, here we are primarily motivated to improve the perceptual quality when applied to natural images. 
Most current single image SR methods use empirical risk minimisation, often with a pixel-wise mean squared error (MSE) loss \citep{dong2014image,shi2016real}. However, MSE, and convex loss functions in general, are known to have limitations when presented with uncertainty in multimodal and nontrivial distributions such as distributions over natural images. In SR, a large number of plausible images can explain the LR input and the Bayes-optimal behaviour for any MSE trained model is to output the mean of the plausible solutions weighted according to their posterior probability. For natural images this averaging behaviour leads to blurry and over-smoothed outputs that generally appear implausible, i.e. the produced estimates have low probability under the natural image prior.
\begin{figure}
  \vspace{-.5cm}
  \begin{minipage}{\textwidth}
  \begin{minipage}[b]{0.69\textwidth}
    \centering
   \scalebox{0}{%
      \begin{tikzpicture}
          \begin{axis}[hide axis]
              \addplot [mark=*, color=teal!70!blue, dashed, line width=0.6pt, forget plot] (0,0);\label{legend:MAP_all}
              \addplot [color=black, dashed, line width=0.6pt, forget plot] (0,0);\label{legend:affinesubspace}
              \addplot [mark=*, color=red,dashed,line width=0.8pt, forget plot] (0,0);\label{legend:AffMSE}
              \addplot [mark=*,color=green!70!black,dashed,line width=0.8pt, forget plot] (0,0);\label{legend:AffMAE}
              \addplot [mark=*,color=teal!70!blue,dashed,line width=0.8pt, forget plot] (0,0);\label{legend:AffGAN}
              \addplot [mark=*,color=violet,dashed,line width=0.8pt, forget plot] (0,0);\label{legend:AffDG}
          \end{axis}
      \end{tikzpicture}%
      }

   \captionsetup{type=figure}
   \includegraphics[width=1.0\linewidth]{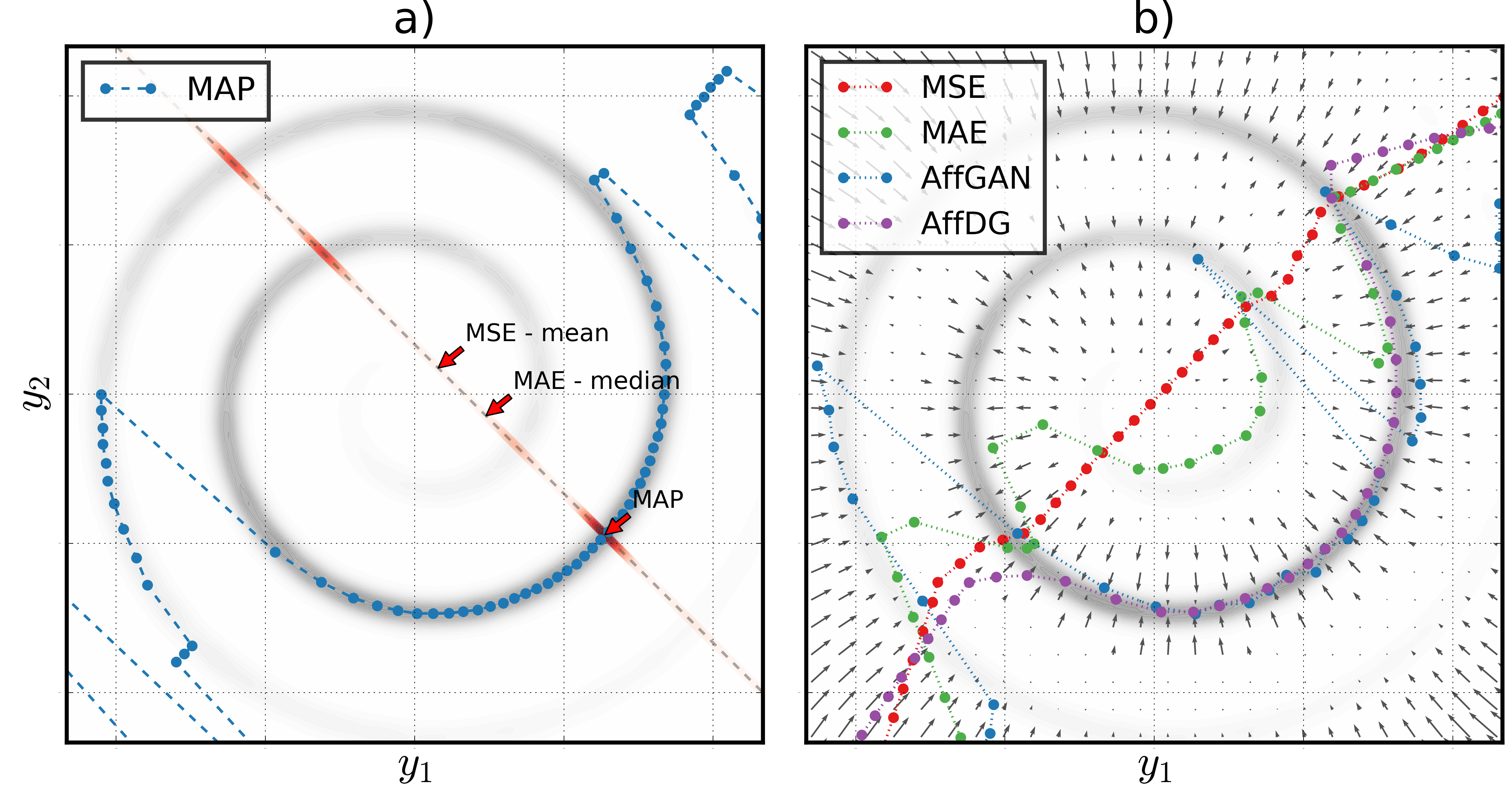}
     \captionof{figure}{Illustration of the SR problem via a toy example. Two-dimensional HR data $y=[y_1,y_2]$ is drawn from a Swiss-roll distribution (in gray). Downsampling is modelled as $x=\frac{y_1+y_2}{2}$. a) Given observation $x=0.5$, valid SR solutions lie along the line $y_2 = 1 - y_1$\,(\ref{legend:affinesubspace}). The red shading illustrates the magnitude of the posterior $p_{Y\vert X=0.5}$. Bayes-optimal estimates under MSE and MAE as well as the MAP estimate given $x=0.5$ are marked with labels. The MAP estimates for different values of $x\in[-8,8]$ are also shown\,(\ref{legend:MAP_all}). b) Trained model outputs for $x\in[-8,8]$ and estimated gradients from a denoising function trained on $p_Y$. Note the \affgan{}(\ref{legend:AffGAN}) and \affdenoiser{}(\ref{legend:AffDG}) models fit the posterior mode well whereas the MSE\,(\ref{legend:AffMSE}) and MAE\,(\ref{legend:AffMAE}) model outputs generally fall in low probability regions.
     }
     \label{fig:swiss_roll_map}
  \end{minipage}
  \hfill
  \begin{minipage}[b]{0.3\textwidth}
    \tiny
    \centering
  \captionsetup{type=table}
  \begin{tabular}{lll}
                        & $\mathbb{H}[q_\theta,p_Y]$ & $\ell_{MSE}(x,Ay)$ \\
    \hline\\
    MAP                 & $3.15$           & - \\
    MSE             & $9.10$    & $1.25\cdot10^{-2}$\\ 
    MAE             & $6.30$    & $4.04\cdot10^{-2}$\\ 
    \affgan{}             & $4.10$    & $0.0$\\             
    \softgan{} & $4.25$    & $8.87\cdot10^{-2}$\\ 
    \affdenoiser{}             & $\bf{3.81}$    & $0.0$\\ 
    \softdenoiser{} & $4.19$    & $1.01\cdot10^{-1}$\\ 

  \hline
  \end{tabular}
      \captionof{table}{Directly estimated cross-entropy $\mathbb{H}[q_\theta,p_Y]$ values. The \affgan{} and \affdenoiser{} achieves cross-entropy values close to the MAP solution confirming that they minimize the desired quantity. The MSE and MAE models performs worse since they do not minimize the cross-entropy. Further the models using affine projections (Aff) performs better than the soft constrained models.}
      \label{tab:swissroll}
    \end{minipage}
  \end{minipage}
  \vspace{-.3cm}
\end{figure}

An idealised method for our applications would use a \emph{full-reference perceptual} loss function that describes the sensitivity of the human visual perception system to different distortions. However the most widely used loss functions MSE and the related peak-signal-to-noise-ratio (PSNR) metric have been shown to correlate poorly with human perception of image quality \citep{laparra2016perceptual,wang2004image}. Improved perceptual quality metrics have been proposed, the most popular being structural similarity (SSIM) \citep{wang2004image} and its multi-scale variants \citep{wang2003multiscale}. 
Although the correlation of these metrics with human perception has improved, they still do not provide a fully satisfactory alternative to MSE for training of neural networks (NN) for SR.

In lieu of a satisfactory perceptual loss function, we leave the empirical risk minimisation framework and present methods based only on natural image statistics. In this paper we argue that a desirable approach is to employ amortised Maximum a Posteriori (MAP) inference, preferring solutions that have a high posterior probability and thus high probability under the image prior while keeping the computational benefits of amortised inference. To motivate why MAP inference is desirable consider the toy problem in Figure \ref{fig:swiss_roll_map}a, where the HR data is  two-dimensional $y=[y_1,y_2]$ and distributed according to the Swiss-roll density. The LR observation is defined as the average of the two pixels $x = \frac{y_1+y_2}{2}$. Consider observing a LR data point $x=0.5$: the set of possible HR solutions is the line $y_1 = 2x-y_2$, more generally an affine subspace, which is shown by the dashed line in Figure \ref{fig:swiss_roll_map}a. The posterior distribution $p(y\vert x)$ is thus degenerate, and corresponds to a slice of the prior along this line, as shown by the red shading. If one minimise MSE or Mean Absolute Error (MAE), the Bayes-optimal solution will lie at the mean or the median along the line, respectively. This example illustrates that MSE and MAE can produce output with very low probability under that data prior whereas MAP inference would always find the mode which by definition is in a high-probability region. See Section \ref{sec:criticism} for a discussion of possible limitations of the MAP inference approach. 

Our first contribution is a convolutional neural networks (CNN) architecture designed to exploit the structure of the SR problem. Image downsampling is a linear transformation, and can be modelled as a strided convolution. As Figure \ref{fig:swiss_roll_map}a illustrates, the set of HR images $y$ that are compatible with any LR image $x$ span an affine subspace. We show that by using specifically chosen linear convolution and deconvolution layers we can implement a projection to this affine subspace. This ensures that our CNNs always output estimates that are consistent with the inputs. The affine projection layer can be added to any CNN, or indeed, any other trainable SR algorithm. Using this architecture we show that training the model for MAP inference reduces to minimising the cross-entropy $\mathbb{H}[q_G,p_Y]$ between the HR data distribution $p_Y$ and the implied distribution $q_G$ of the model's output when evaluated at random LR images. As a result, we don't need corresponding HR and LR image pairs any more, and training becomes more akin to training generative models. However direct minimisation of the cross-entropy is not possible and instead we develop three approaches, all depending on projecting the model output to the affine subspace of valid solution, to approximate it directly from data: %
\begin{enumerate}
\item We present a variant of the Generative Adversarial Networks (GAN) \citep{goodfellow2014generative} which approximately minimises the Kullback–Leibler divergence ($\operatorname{KL}$) and cross-entropy between $q_G$ and $p_Y$. Our analysis provides theoretical grounding for using GANs in image SR \citep{ledig2016photo}. We also introduce a trick that we call \emph{instance noise} that can be generally applied to address the instability of training GANs.
\item We employ denoising as a way to capture natural image statistics. Bayes-optimal denoising approximately learn to take a gradient step along the log-probability of the data distribution \citep{alain2014regularized}. These gradient estimates from denoising can be directly backpropagated through the network to minimise cross-entropy between $q_G$ and $p_Y$ via gradient descent.
\item We present an approach where the probability density of data is directly modelled via a generative model trained by maximum likelihood. We use a differentiable generative model based on PixelCNNs \citep{van2016pixel} and Mixture of Conditional Gaussian Scale Mixtures \citep[MCGSM,][]{Theis2012a} whose performance we believe is very close to the-state-of-the-art in this category.
\end{enumerate}
In section \ref{sec:results} we empirically demonstrate the behaviour of the proposed methods on both the two dimensional toy dataset and on real image datasets. Lastly, in Appendix \ref{sec:appendix_variational_affgan}  we show that a stochastic version of AffGAN performs amortised variational inference, which for the first time establishes a connection between GANs and variational inference as in e.\,g.\ variational autoencoders \citep{kingma2014variational}.

\section{Related work}
The GAN framework was introduced by \cite{goodfellow2014generative} which also showed that these models minimise the Shannon-Jensen Divergence between $q_G$ and $p_Y$ under certain conditions. In Section \ref{sec:GAN}, we present an update rule that corresponds to minising $\operatorname{KL}[q_G \| p_Y]$. Recently, \citet{nowozin2016f} presented a more general treatment that connects GANs to $f$-divergence minimisation. In parallel to our contributions, theoretical work by \citet{Shakir2016} presented a unifying view on learning in GAN-style algorithms, of which our variant can be regarded a special case. The focus of several recent papers on GANs were algorithmic tricks to improve their stability \citep{radford2015unsupervised,salimans2016improved}. In Section \ref{sec:instancenoise} we introduce another such trick we call \emph{instance noise}. We discuss theoretical motivations for this and compare it to \emph{one-sided label smoothing} proposed by \cite{salimans2016improved}. We also refer to parallel work by \citet{arjovsky2017towards} proposing a similar method. Recently, several attempts have been made to improve perceptual quality of SR using deep representations of natural images. \citet{bruna2015super} and \citet{li2016markov} measure the Euclidean distance in the nonlinear feature space of a deep NN pre-trained to perform object classification.  \citet{dosovitskiy2016generating} and \citet{ledig2016photo} use a similar approach and also add an adversarial loss term. Unpublished work by \citet{garcia2016github} explored combining GANs with an $L_1$ penalty between the LR input and the down-sampled output. We note that the soft $L_2$ or $L_1$ penalties used in these methods can be interpreted as assuming Gaussian and Laplace observation noise. In contrast, our approach assumes no observation noise and satisfies the consistency of inputs and outputs exactly by using an affine projection as explained in Section \ref{sec:theory_likelihood}. In other work, \citet{larsen2015autoencoding} proposed to replace the pixel-wise MSE used for training of variational autoencoders with a learned metric from the GAN discriminator. Our denoiser based method exploits a fundamental connection between probabilistic modelling and learning to denoise \citep[see e.\,g.\ ][]{vincent2008extracting,alain2014regularized,sarela2005denoising,rasmus2015semi,greff2016tagger}: a Bayes-optimal denoiser can be used to estimate the gradient of the log probability of data. To our knowledge this work is the first time that the output of a denoiser is explicitly back-propagated to train another network. Lastly, we note that denoising has been used to solve inverse problems in compressed sensing as in approximate message passing \citep{metzler2015AMP}.
\section{Theory\label{sec:theory}}
Consider a function $f_{\theta}(x)$ parametrised by $\theta$ which maps a LR observation $x$ to a HR estimate $\hat{y}$. Most current SR methods optimise model parameters via empirical risk minimization:
\begin{align}
    \argmin_{\theta} \mathbb{E}_{y,x}[\ell(y,f_{\theta}(x))]
\end{align}
Where $y$ is the true target and $\ell$ is some loss function. The loss function is typically a simple convex function most often MSE $\ell_{\textrm{MSE}}(y,\hat{y}) = \left\|y - \hat{y}\right\|^2_2$ as in \citep{dong2014image,shi2016real}. Here, we seek to perform MAP inference instead. For a single LR observation the MAP estimate is
\begin{align}
  \hat{y}(x) = \argmax_y \log p_{Y\vert X}(y \vert x)
\end{align}
Instead of calculating $\hat{y}$ for each $x$ separately we perform amortised inference, i.\,e.\ we would like to train the SR function $f_\theta(x)$ to calculate the MAP estimate. A natural loss function for learning the parameters $\theta$ is the average log-posterior:
\begin{align}
  \argmax_{\theta} \mathbb{E}_{x}\log p_{Y\vert X}(f_{\theta}(x)\vert x),
\end{align}
where the expectation is taken over the distribution of LR observations $x$. This loss depends on the unknown posterior distribution $p_{Y|X}$. We proceed by decomposing the log-posterior using Bayes' rule as follows.
\begin{align}
\argmax_{\theta} \left\{ \underbrace{\mathbb{E}_{x} \log p_{X\vert Y}(x \vert f_\theta(x))}_{\text{Likelihood}} + \underbrace{\mathbb{E}_{x}\log p_{Y}(f_\theta(x))}_{\text{Prior}} - \underbrace{\mathbb{E}_{x}\log p_{X}(x)}_{\text{Marginal Likelihood}} \right\}.
  \label{eq:map_objective_expandend}
\end{align}

\subsection{Handling the Likelihood term\label{sec:theory_likelihood}}
Notice that the last term of Eqn.\ \eqref{eq:map_objective_expandend}, the marginal likelihood, does not depend on $\theta$, so we only have to deal with the likelihood and image prior. The observation model in SR can be described as follows.
\begin{equation}
  x = A \hat{y},
  \label{eq:io_consistency}
\end{equation}
where $A$ is a linear transformation used for image downsampling. In general, $A$ can be modelled as a strided two-dimensional convolution. Therefore, the likelihood term in Eqn. \eqref{eq:map_objective_expandend} is degenerate $p(x \vert f_{\theta}(x)) = \delta(x-Af_{\theta}(x))$, and Eqn. \eqref{eq:map_objective_expandend} can be rewritten as constrained optimisation:
\begin{align}
  \argmax_{\substack{\theta\\\forall x: Af_{\theta}(x)=x}}  \mathbb{E}_{x}[\log p_Y(f_{\theta}(x))]\label{eq:MAP_objective_constrained}
\end{align}

To satisfy the constraints, we introduce a parametric function class that always guarantees $Af_{\theta}(x)=x$. Specifically, we propose to use functions of the form
\begin{equation}
  g_\theta(x) = \Pi^{A}_xf_\theta(x) = (I-A^+A)f_{\theta}(x) + A^+x
\end{equation}
where $f_{\theta}$ is an arbitrary mapping from LR to HR space, $\Pi^{A}_x$ a projection to the affine subspace $\{y: yA = x\}$, and $A^+$ is the Moore-Penrose pseudoinverse of $A$, which satisfies $AA^+A = A$ and $A^+AA^+ = A^+$. Conveniently, if $A$ is a strided two-dimensional convolution, then $A^+$ becomes a deconvolution or up-convolution, which is a standard operation used in deep learning \citep[e.\,g.\ ][]{shi2016real}. It is important to stress that the optimal deconvolution $A^+$ is not simply the transpose of $A$, Figure \ref{fig:mse_celeba} illustrates the upsampling kernel ($A^{+}$) that corresponds to a Gaussian downsampling kernel ($A$). For any $A$ the deconvolution $A^+$ can be easily found, here we used numerical methods as detailed in Appendix\ \ref{sec:app_affproj}. Intuitively, $A^+x$ can be thought of as a baseline SR solution, while $(I-A^+A)f_\theta$ is the residual. The operation $(I-A^+A)$ is a projection to the null-space of $A$, therefore when we downsample the residual $(I-A^+A)f_{\theta}$ we are guaranteed to get $0$ no matter what $f_{\theta}$ is. By using functions of this form we can turn Eqn.\ \eqref{eq:MAP_objective_constrained} into an unconstrained optimization problem.
\begin{equation}
 \argmax_{\theta} \mathbb{E}_{x} \log p_Y(\Pi^{A}_x f_{\theta}(x))
\end{equation}

Interestingly, the objective above can be expressed in terms of the probability distribution of the model output $q_{\theta}(y) := \int \delta\left(y-\Pi^{A}_x f_{\theta} (x)\right) p_{X}(x) dx$ as follows.
\begin{equation}
 \argmax_{\theta} \mathbb{E}_{x} \log p_Y(\Pi^{A}_x f_{\theta}(x)) = \argmax_\theta \mathbb{E}_{\hat{y} \sim q_\theta} \log p_Y(\hat{y}) = \argmin_\theta \mathbb{H}[q_\theta, p_Y],\label{eq:MAP_objective_crossentropy}
\end{equation}
where $\mathbb{H}[q,p]$ denotes the cross-entropy between $q$ and $p$ and we used $\mathbb{H}[q_\theta,p_Y]=\mathbb{E}_{\hat{y} \sim q_\theta} \left[-\log p_Y(\hat{y})\right]$. To minimise this objective, we do not need matched input-output pairs as in empirical risk minimisation. Instead we need to match the marginal distribution of reconstructed images $q_\theta$ to that of the distribution of HR images. In this respect, the problem becomes more akin to unsupervised learning or generative modelling. In the following sections we present three approaches to finding the optimal $\theta$ utilising the properties of the affine projection.

\subsection{Affine projected Generative Adversarial Networks \label{sec:GAN}}

Generative Adversarial Networks \citep{goodfellow2014generative} consist of a generator $G$ that turns noise sampled from some distribution $z\sim p_Z$ into images $G(z)$ via a parametric mapping, and a discriminator $D$ that learns to distinguish between real and synthetic images. The generator and discriminator are updated in tandem resulting in the generative distribution $q_G$ moving closer to the distribution of real data $p_Y$. The behaviour of GANs depends on the specifics of how the generator and the discriminator are trained. We use the following objective functions for $D$ and $G$:
\begin{align}
  \mathcal{L}(D; G) &= - \mathbb{E}_{y\sim p_Y} \log D(y) - \mathbb{E}_{z\sim p_Z}\log(1-D(G(z))\label{eq:GANupdaterule}, \\
  \mathcal{L}(G; D) &= - \mathbb{E}_{z\sim p_Z} \log \frac{D(G(z))}{1-D(G(z))}\nonumber.
\end{align}
The algorithm iterates two steps: first, it updates $D$ by lowering $\mathcal{L}(D; G)$ keeping $G$ fixed, then it updates $G$ by lowering $\mathcal{L}(G; D)$ keeping $D$ fixed. It can be shown that this amounts to minimising $\operatorname{KL}[q_G\|p_Y]$, where $q_G$ is the distribution of samples generated by $G$. See Appendix \ref{sec:app_GAN_convergence} for a proof\footnote{First shown in \citep{huszar2016alternative}.} In the context of SR, the affine projected SR function $\Pi^A_x f_\theta$ takes the role of the generator. Instead of noise, the generator is now fed low-resolution images $x\sim p_X$. Leaving everything else unchanged, we can deploy the GAN algorithm to minimise $\operatorname{KL}[q_\theta \| p_Y]$. We call this algorithm \emph{affine projected GAN} or \affgan{} for short. Similarly, we introduce notation \softgan{} to denote the GAN algorithm without the affine projection, which instead uses an additional soft-constraint $\ell_{LR}=\textrm{MAE}(x,A\hat{y})$ as in \citep{garcia2016github}. Note that the difference between the cross-entropy and the KL divergence is the entropy of $q_{\theta}$: $\mathbb{H}[q_{\theta},p_Y] - \operatorname{KL}[q_{\theta}\|p_Y] = \mathbb{H}[q_{\theta}]$. Hence, we can expect \affgan{} to favour approximate MAP solutions that lead to higher entropy and thus more diverse solutions overall.

\subsubsection{Instance Noise \label{sec:instancenoise}}
The theory suggests that GANs should be a convergent algorithm. If a unique optimal discriminator exists and it is reached by optimising $D$ to perfection at each step, technically the whole algorithm corresponds to gradient descent on an estimate of $\operatorname{KL}[q_\theta \| p_Y]$ with respect to $\theta$. In practice, however, GANs tend to be highly unstable. So where does the theory go wrong? We think the main reason for the instability of GANs stems from $q_{\theta}$ and $p_Y$ being concentrated distributions whose support does not overlap. The distribution of natural images $p_Y$ is often assumed to concentrate on or around a low-dimensional manifold. In most cases, $q_\theta$ is degenerate and manifold-like by construction, such as in \affgan{}. Therefore, odds are that especially before convergence is reached, $q_{\theta}$ and $p_Y$ can be perfectly separated by several $D$s violating a condition for the convergence proof. We try to remedy this problem by adding \emph{instance noise} to both SR and true image samples. This amounts to minimising the divergence $d_\sigma(q_{\theta},p_Y) = \operatorname{KL}\left[p_{\sigma} \ast q_\theta \middle\| p_{\sigma} \ast p_Y\right]$, where $p_{\sigma}\ast q_{\theta}$ denotes convolution of $q_{\theta}$ with the noise distribution $p_{\sigma}$. The noise level $\sigma$ can be annealed during training, and the noise allows us to safely optimise $D$ until convergence in each iteration. The trick is related to one-sided label noise introduced by \citet{salimans2016improved}, however without introducing a bias in the optimal discriminator, and we believe it is a promising technique for stabilising GAN training in general. For more details please see Appendix \ref{appdx:instancenoise}

\subsection{Denoiser Guided Super-Resolution}

To optimise the criterion Eqn. \eqref{eq:MAP_objective_constrained} via gradient descent we need its gradient with respect to $\theta$:
\begin{align}
 \frac{\partial}{\partial \theta}\mathbb{E}_{x}[\log p(\Pi^A_xf_\theta(x))] = \mathbb{E}_{x} \left[ \left.\frac{\partial}{\partial y} \log p(y)\right|_{y=\Pi^A_xf_\theta(x)} \cdot \Pi^A_x\frac{\partial}{\partial \theta} f_\theta(x) \right]\label{eq:logPgrad}
\end{align}
Here $\frac{\partial}{\partial \theta} f_\theta$ are the gradients of the SR function which can be calculated via back-propagation whereas $\frac{\partial}{\partial y} \log p_Y(y)$ requires estimation since $p_Y$ is unknown. We use results from \citep{alain2014regularized,sarela2005denoising} showing that in the limit of infinitesimal Gaussian noise, optimal denoising functions can be used to estimate this gradient:
\begin{align}
  f^\ast_\sigma = \argmin_f \mathbb{E}_{y\sim p_Y} \ell_{MSE}(f(y + \sigma\epsilon), y) \implies \frac{f^\ast(y) - y}{\sigma^2} \approx \frac{\partial}{\partial y} \log p_Y(y),\label{eq:DAE_gradestimate}
\end{align}
where $\epsilon\sim \mathcal{N}(0,I)$ is Gaussian white noise, $f^\ast_{\sigma}$ is the Bayes-optimal denoising function for noise level $\sigma$. Using these results we can maximise Eqn. \eqref{eq:MAP_objective_crossentropy} by first training a neural network to denoise samples from $p_Y$ and then backpropagate the gradient estimates from Eqn. \eqref{eq:DAE_gradestimate} via the chain rule in Eqn. \eqref{eq:logPgrad} to update $\theta$. Well call this method \affdenoiser{}, as it uses the affine subspace projection and is guided by the gradient from the DAE. Similar to above we'll call the similar algorithm soft-enforcing Eqn. \eqref{eq:io_consistency} \softdenoiser{}.

\subsection{Density Guided Super-Resolution}

As a more direct baseline model for amortised MAP inference we fit a tractable, yet powerful density model to $p_Y$ using maximum likelihood, and then use cross entropy with respect to the generative model to approximate Eqn. \eqref{eq:MAP_objective_crossentropy}. We use a deep generative model similar to the pixelCNN \citep{van2016pixel} but with a continuous (and differentiable) MCGSM \citep{Theis2012a} likelihood. These type of models are state-of-the-art in density estimation, are relatively fast to evaluate and produce visually interesting samples \citep{van2016pixel}. We call this method \affdensity{}, as it uses the affine projection and is guided by the log-likelihood of a density model.

\section{Experiments}

We designed our experiments to address the following questions:
\begin{itemize}
  \item Are the methods proposed in Section\,\ref{sec:theory} successful at minimising cross-entropy?\,$\rightarrow$\,Section \,\ref{sec:results_swissroll}
  \item Does the affine projection layer hurt the performance of CNNs for image SR?\,$\rightarrow$\,Section\,\ref{sec:results_mse_celebA}
  \item Do the proposed methods produce perceptually superior SR results? $\rightarrow$\,Sections\,\ref{sec:results_grass}-\ref{sec:results_imagenet}
\end{itemize}

We initially illustrate the behaviour of the proposed algorithms on data where exact MAP inference is computationally tractable. Here the HR data $y=[y_1,y_2]$ is drawn from a two-dimensional noisy Swiss-roll distribution and the one-dimensional LR data $x$ is simply the average of the two HR pixels. Next we tested the proposed algorithm in a series of experiments on natural images using $4\times$ downsampling.. For the first dataset, we took random crops from HR images containing grass texture. SR of random textures is known to be very hard using MSE or MAE loss functions. Finally, we tested the proposed models on real image data of faces\,(\texttt{Celeb-A}) and natural images\,(\texttt{ImageNet}). All models were convolution neural networks implemented using Theano \citep{team2016theano} and Lasagne \citep{lasagne}. We refer to Appendix \ref{appdx:experimental_details} for full experimental details. 

\section{Results and Discussion\label{sec:results}}

\subsection{2D MAP inference: Swiss-Roll}
\label{sec:results_swissroll}

In this experiment we wanted to demonstrate that \affgan{} and \affdenoiser{} are indeed minimising the MAP objective in Eqn. \eqref{eq:MAP_objective_crossentropy}. For this we used the two-dimensional toy problem where $p_Y$ can be evaluated using brute-force Monte Carlo. Figure \ref{fig:swiss_roll_map}b) shows the outputs for $x=[-8,8]$ for models trained with different criterion. The \affgan{} and \affdenoiser{} solutions largely fit the dominant mode similar to MAP inference. For the MSE and MAE models the output generally falls in regions with low prior density. Table \ref{tab:swissroll} shows the cross-entropy $\mathbb{H}[q_\theta,p_Y]$ achieved by different methods, averaged over 10 independent trials with random initialisation. The cross-entropy values for the GAN and DAE based models are relatively close to the optimal MAP solution, which in this case we can find in a brute-force way. As expected the MSE and MAE models perform worse as these models do not minimize $\mathbb{H}[q_\theta,p_Y]$. We also calculated the average MSE between the network input and the downsampled network output. For the affine projected models, this error is exactly $0$. The soft constrained models only approximately satisfy this constraint, even after extensive training (Table \ref{tab:swissroll} second column). Further, we observe that the affine projected models generally found a lower cross-entropy $\mathbb{H}[q_\theta,p_Y]$ when compared to soft-constrained versions.

\subsection{Affine Projected Networks: Proof of Concept using MSE criterion}
\label{sec:results_mse_celebA}

\definecolor{red}{RGB}{228,26,28}
\definecolor{blue}{RGB}{55,126,184}
\definecolor{green}{RGB}{77,175,74}
\definecolor{purple}{RGB}{152,78,163}

\begin{figure}
  \center
  \scalebox{0}{%
     \begin{tikzpicture}
         \begin{axis}[hide axis]
             \addplot [color=red, line width=1.0pt, forget plot] (0,0);\label{legend:MSEFT}
             \addplot [color=blue, line width=1.0pt, forget plot] (0,0);\label{legend:MSETT}
             \addplot [color=purple, line width=1.0pt, forget plot] (0,0);\label{legend:MSETR}
             \addplot [color=green, line width=1.0pt, forget plot] (0,0);\label{legend:MSE}
         \end{axis}
     \end{tikzpicture}
     }

  \includegraphics[width=1.0\linewidth]{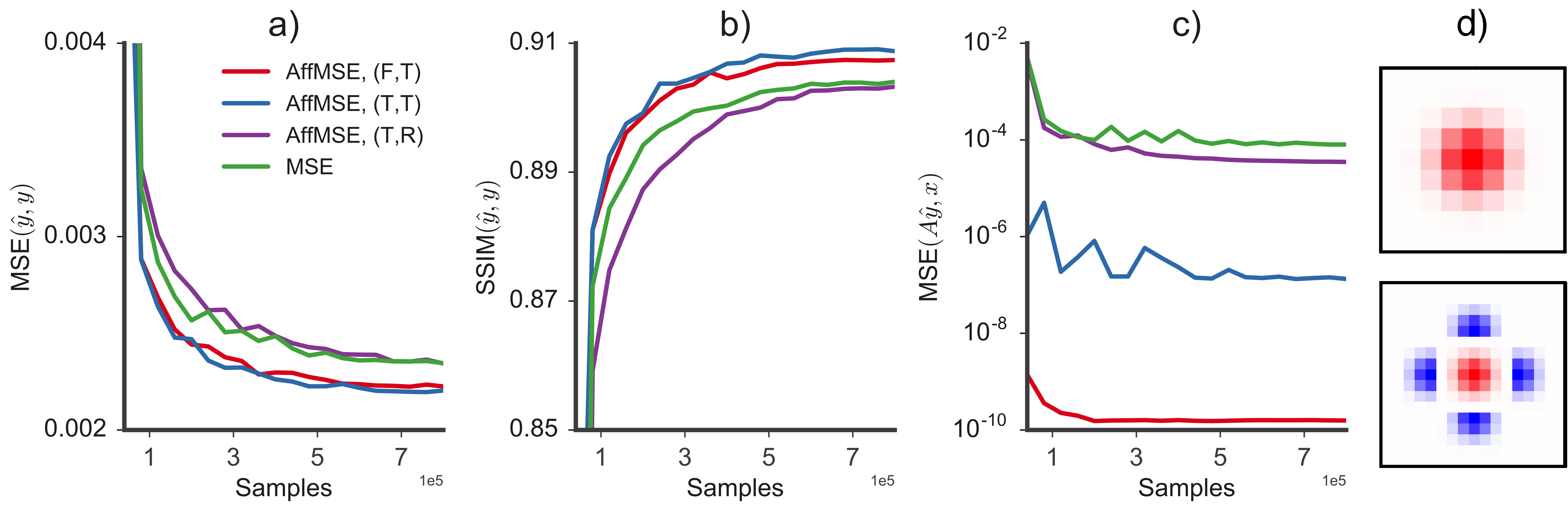}
  \caption{CelebA performance for MSE models during training. The distance between HR model output $\hat{y}$ and true HR image $y$ using MSE in a) and SSIM in b). MSE in LR space between input $x$ and down-sampled model output $A\hat{y}$ in c). The tuple in the legend indicate: (\textbf{(F)}ixed  / \textbf{(T)}rainable affine projection, \textbf{(T)}rained / \textbf{(R)}andom initialised affine projections). The models using pre-trained affine projections (fixed:\ref{legend:MSEFT}, trainable:\ref{legend:MSETT}) always performs better in all metrics compared to models using either random initialized affine projections (\ref{legend:MSETR}) or no projection (\ref{legend:MSE}). Further, a fixed pre-trained affine projection ensures the best consistency between input and down-sampled output as seen in figure c). $A$ (top) and $A^+$ (bottom) kernels of the affine projection are seen in d). }
  \label{fig:mse_celeba}
\end{figure}

Adding the affine projection $\Pi^{A}_x$ restricts the class of functions that the SR network can model, so it is important to verify that the network is still capable of achieving the same performance in SR as unconstrained CNN architectures. To test this, we trained CNNs with and without affine projections to perform SR on the \texttt{CelebA} dataset using MSE as the objective function. Results are shown in Figure~\ref{fig:mse_celeba}. First note that when using affine projections, a randomly initialised network starts learning from a lower initial loss as the low-frequency components of the network output already match those of the target image. We observed that the affine projected networks generally train faster than unconstrained ones. Furthermore, the affine projected networks tend to find a better solution as measured by MSE and SSIM (Figure~\ref{fig:mse_celeba}a-b). To investigate which aspects of the network architecture are responsible for the improved performance, we evaluated two further models: In one variant, we initialise the affine projected CNN to implement the correct projection, but then treat $A^{+}$ as a trainable parameter. In the final variant, we keep the architecture the same, but initialise the final deconvolution layer $A^{+}$ randomly and allow it to be trained. We found that initialising $A^{+}$ to the correct Moore-Penrose inverse is important, and we get the similar results irrespective of whether or not it is fixed during training. Figure~\ref{fig:mse_celeba}c shows the error between the network input and the downsampled network output. We can see that the exact affine projected network keeps this error at virtually $0.0$ (up to numerical precision), whereas any other network will violate this consistency. In Figure~\ref{fig:mse_celeba}d we show the downsampling kernel $A$ and the corresponding optimal kernel for $A^+$.

\subsection{Grass Textures}
\label{sec:results_grass}
\begin{figure}
  \center
  \includegraphics[width=0.9\linewidth]{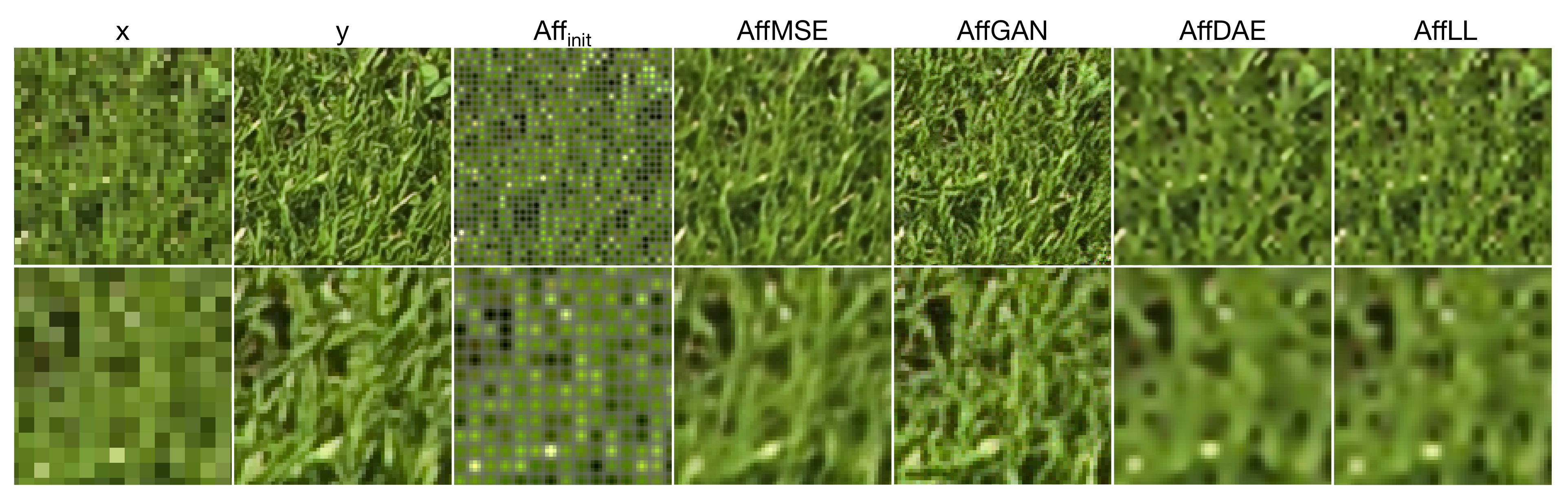}
  \caption{4$\times$ SR of grass textures. Top row shows LR model input $x$, true HR image $y$ and model outputs according to figure legend. Bottom row shows zoom in on except from the images in the top row. The \affgan{} image is much sharper than the somewhat blurry \affmse{} image. Note that both the \affdenoiser{} and \affdensity{} produces very blurry results. The $\text{Aff}_{\text{init}}$ shows the output from an untrained affine projected model, i.\,e.\ the baseline solution, illustrating the effect of the upsampling using $A^+$.}
  \label{fig:grass}
\end{figure}

Random textures are known to be hard model using MSE loss function. Figure~\ref{fig:grass} shows $4\times$ SR of grass texture patches using identical affine projected CNNs trained with different loss functions. When randomly initialised, affine projected CNNs always produce an output with the correct low-frequency components,as illustrated by the third panel labelled $\text{Aff}_\text{init}$ in Figure~\ref{fig:grass}. The \affgan{} model produces clearly the sharpest images, and we found the images to be plausible given the LR inputs. Notice that the reconstruction is not perfect pixel-by-pixel, but it has the correct statistical properties for the human visual system to recognise it as grass texture. The \affdenoiser{} and \affdensity{} models both produced blurry results which we where unable to improve upon using various optimization methods. Due to these findings we choose not to perform any further experiments with these models and concentrate on \affgan{} instead. We refer to Appendix \ref{appdx:DG_models} for discussion of the results of these models.

\subsection{CelebA Faces}
\label{sec:results_celebA}

\begin{figure}[!ht]
  \begin{minipage}{\textwidth}
  \begin{minipage}[b]{0.69\textwidth}
    \centering
   \captionsetup{type=figure}
    \includegraphics[width=1.0\linewidth]{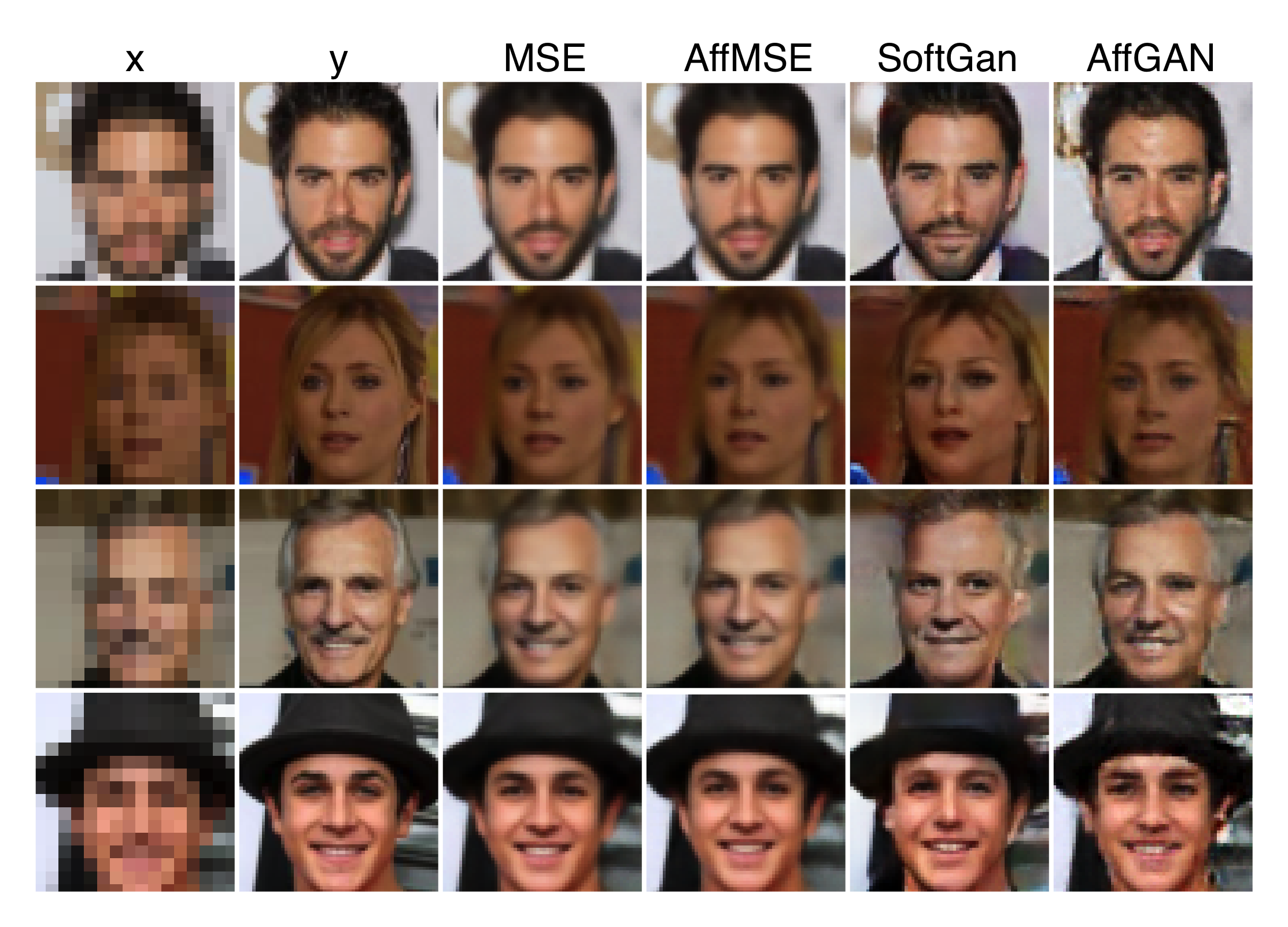}
    \caption{4$\times$ SR of CelebA faces. Model input $x$, target $y$ and model outputs according to figure legend. Both the \affgan{} and \softgan{} produces clearly shaper images than the blurry MSE outputs. We found that \affgan{} outputs slightly sharper images compared to \softgan{}, however also with slightly more high-frequency noise.}
    \label{fig:celeba_panel}
  \end{minipage}
  \hfill
  \begin{minipage}[b]{0.3\textwidth}
    \tiny
    \centering
  \captionsetup{type=table}
  \tiny
    \begin{tabular}{llll}
          & SSIM   & PSNR     & $\ell_{MSE}(x,A\hat{y})$\\
    \hline\\
    MSE & $0.90$ & $26.30$  & $8.0\cdot10^{-5}$\\
    \affmse{} & $0.91$ & $26.53$  & $1.6\cdot10^{-10}$\\
    \softgan{} & $0.76$ & $21.11$  & $2.3\cdot10^{-3}$\\
    \affgan{} & $0.81$ & $23.02$  & $9.1\cdot10^{-10}$\\
    \hline
  \end{tabular}
  \caption{PSNR, SSIM and MSE scores for the CelebA dataset. In terms of PSNR and SSIM in HR space the MSE trained models achieves the best scores as expected and the \affgan{} performs better than the \softgan{}. Considering $\ell_{MSE}(x,A\hat{y})$ the models using the affine projections (Aff) clearly show better consistency between input $x$ and down sampled model output $A\hat{y}$ than models not using the projection.}
  \label{tab:celeba}
    \end{minipage}
  \end{minipage}
\end{figure}

In Figure~\ref{fig:celeba_panel} the SR results are seen for several models trained using different loss functions. The MSE trained models outputs somewhat generic and over-smoothed images as expected. For the GAN models the global content is correct for both the affine projected and soft constrained models. Comparing the \affgan{} and \softgan{} outputs the \affgan{} model produces slightly sharper images which however also seem to contain slightly more high frequency noise. We observed some colour drifting for the soft constrained models. Table \ref{tab:celeba} shows quantitative results for the same four models where, in terms of PSNR and SSIM, the MSE model achieves the best scores as expected. The consistency between input and output clearly shows that the models using the affine projections satisfy Eqn.\ \eqref{eq:io_consistency} better than the soft constrained versions for both MSE and GAN losses.

\subsection{Natural Images}
\label{sec:results_imagenet}

\begin{figure}
  \center
  \includegraphics[width=0.8\linewidth]{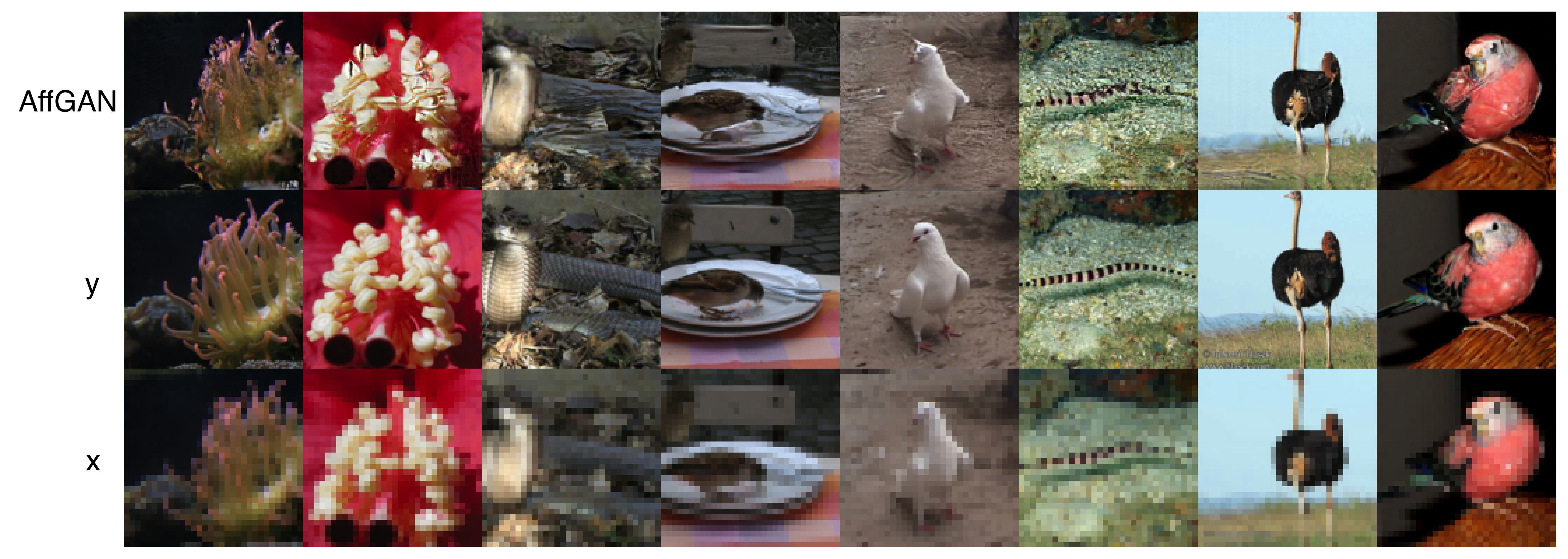}
  \caption{$4\times$ SR from $32\times32$ to $128\times128$ using \affgan{} on the ImageNET. \affgan{} outputs (top row), true HR images $y$ (middle row), model input $x$ (bottom row). Generally the \affgan{} produces plausible outputs which are however still easily distinguishable from true images. Interestingly the snake depicted in the third column is super resolved into water which is obviously wrong but still a very plausible image considering the LR input image.}
  \label{fig:imagenet_panel1}
\end{figure}

In Figure~\ref{fig:imagenet_panel1} we show the results for $4\times$ SR from $32\times32$ to $128\times128$ pixels for \affgan{} trained on natural images from ImageNET. For most of the images the results are sharp and corresponds well with the LR input. However we still see the high-frequency noise present in most GAN results in some of the images. Interestingly the snake depicted in the third column is super resolved into water which is obviously wrong but still a very plausible image considering the LR input image. Further, water will likely have a higher density under the image prior than snakes which suggests that the GAN model dreams up reasonable data.

\subsection{Criticism and future directions\label{sec:criticism}}

One argument against MAP inference is that the mode of a distribution is dependent on the representation: transforming a variable through an invertible transformation and performing MAP inference in the transformed space may lead to different answers depending on the transformation. As an extreme example, consider transforming a continuous random scalar $Y$ with its cumulative distribution function $F = \mathbb{P}(Y\leq\cdot)$. The resulting variable $F(Y)$ is uniformly distributed, so any value in the interval $(0,1]$ can be the mode. Thus, the MAP estimate is not unique if one allows for alternative representations, and there is no guarantee that the MAP estimate in 24-bit RGB pixel representation which we seek in this paper is in any way special. One may arrive at a different solution when performing MAP estimation in the feature space of a convolutional neural network, or even if merely an alternative colour space is used. Interestingly, \affgan{} is more resilient to coordinate transformations: Eqn.\ \eqref{eq:GANupdaterule} includes the extra term $\mathbb{H}[q_\theta]$ which is effected by transformations the same way as $\mathbb{H}[q_\theta,p_Y]$. The second argument relates to the assumption that MAP estimates appear \emph{plausible}. Although by definition the mode lies in a high-probability region, it does not guarantee that its appearance is anything like that of a random sample. Consider for example data drawn from a $d$-dimensional standard Normal distribution. Due to concentration of measure, as $d$ increases the norm of a typical sample will be approximately $\sqrt{d}$ with very high probability. The mode, however, has a norm of $0$. In this sense, the mode of the distribution is highly atypical. Indeed human observers can easily tell apart a typical sample from the noise distribution and the mode, but would have a hard time noticing the difference between two random samples. This argument suggests that sampling from the posterior $p_{Y|X}$ may be a good or even preferable way to obtain plausible reconstructions. In Appendix \ref{sec:appendix_variational_affgan} we establish a connection between variational inference, such as in varational autoencoders \citep{kingma2014variational}, and a stochastic version of \affgan{}, however leaving emperical studies as further.

\section{Conclusion}

In this work we developed methods for approximate MAP inference in SR. We first introduced an architectural restriction to neural networks projecting the model output to the affine subspace of valid solutions. We then proposed three methods, based on GANs, denoising or density models, for amortised MAP inference in SR using this affine projection. In high dimensions we empirically found that the GAN based approach, \affgan{} produced the most visually appealing results. Our work follows successful demonstrations of GAN-based algorithms for image SR \citep{ledig2016photo}, and we provide additional theoretical motivation for why this approach makes sense. In future work we plan to focus on a stochastic extension of \affgan{} which can be seen as performing amortised variational inference.

\small{
\bibliography{iclr2017_conference}
\bibliographystyle{iclr2017_conference}
}

\clearpage

\appendix

\section{Generative Adversarial Networks for minimising KL-divergence}
\label{sec:app_GAN_convergence}
First note that for a fixed generator $G$ the discriminator $D$ maximises:
\begin{align}
  &\mathbb{E}_{y\sim p_Y} \log D_{\psi}(y) + \mathbb{E}_{z\sim \mathcal{N}}\log(1-D_{\psi}(G_{\theta}(z)) = \\
  &\mathbb{E}_{y\sim p_Y} \log D_{\psi}(y) + \mathbb{E}_{y\sim q_G} \left[\log(1-D_{\psi}(y))\right] = \\
  &\int_y p_Y(y)\log D_{\psi}(y) +  q_G(y)\log(1-D_{\psi}(y))dy
\end{align}

where $q_G$ is the generative distribution. A function of the form $a\log(x) + b\log(1-x)$ always has maximum at $\frac{a}{a+b}$ and we find the Bayes-optimal discriminator to be (assuming equal prior class probabilities)

\begin{align}
  D^{*}(y) = \frac{p_Y(y)}{p_Y(y) + q_G(y)}
\end{align}

Let's assume that this Bayes-optimal discriminator is unique and can be approximated closely by our neural network (see Appendix \ref{appdx:instancenoise} for more discussion on this assumption).

Using the modified update rule proposed here the combined optimization problem for the discriminator and generator is
\begin{align}
  V(\psi,\theta) = \max_{\psi,\theta} \mathbb{E}_{y\sim p_Y} \log D_{\psi}(y) + \mathbb{E}_{z\sim \mathcal{N}}\left[\log(D_{\psi}(G_{\theta}(z))-\log(1-D_{\psi}(G_{\theta}(z)) \right] \label{eq:GAN_minmax}
\end{align}

Starting from the definition of $\operatorname{KL}[q_G|p_Y]$

\begin{align}
  \operatorname{KL}[q_G|p_Y] &= \mathbb{E}_{y\sim q_G} \log\frac{q_G(y)}{p_Y(y)}\\
                          &= \mathbb{E}_{y\sim q_G} \log\frac{1-D^*(y)}{D^*(y)} \ \ \text{(insert Bayes-optimal classifier)}\\
                          &\approx \mathbb{E}_{y\sim q_G} \log\frac{1-D_{\psi}(y)}{D_{\psi}(y)} = - \mathbb{E}_{y\sim q_G} \log\frac{D_{\psi}(y)}{1-D_{\psi}(y)}
\end{align}

Which is equal to the terms affecting the generator in Eqn.\ \eqref{eq:GAN_minmax}.

\section{Affine projection}
\label{sec:app_affproj}

\subsection{Numerical Estimation of the pseudo-inverse}

In practice we implement the down-sampling projection $A$ as a strided convolution with a fixed Gaussian smoothing kernel where the stride corresponds to the down-sampling factor. $A^+$ is implemented as a transposed convolution operation with parameters optimised numerically via stochastic gradient descent on the following objective function:

 \begin{align}
  \ell_{1}(B) &= \mathbb{E}_{y\sim \mathcal{N}_rd}\left\| Ay - ABAy \right\|_2^2\\
  \ell_{2}(B) &= \mathbb{E}_{x\sim \mathcal{N}_r}\left\| Bx - BABx \right\|_2^2\\
  A^+ &= \argmin_B \left\{ \ell_1(B) + \ell_{2}(B) \right\}
 \end{align}

Where $\mathcal{N}_d$ is the $d$-dimensional standard normal distribution, and $d$ is the dimensionality of LR data $x$. $\ell_1$ and $\ell_2$ can be thought of as a Monte Carlo estimate of the spectral norm of the transformations $A - ABA$ and $B - BAB$, respectively. The Monte Carlo formulation above has the advantage that it can be optimised via stochastic gradient descent.
The operation $ABA$ can be thought of as a three-layer fully linear convolutional neural network, where $A$ corresponds to a strided convolution with fixed kernels, while $B$ is a trainable deconvolution. We note that for certain downsampling kernels $A$ the exact $A^+$ would have an infinitely large kernel, although it can always be approximated with a local kernel. At convergence we found $\ell_1+\ell_2$ to be between $10^{-12}$ and  $10^{-8}$ depending on the down-sampling factor, width of the Gaussian kernel used for $A$ and the filter sizes of $A$ and $B$.

\subsection{Gradients}

The gradients of the affine projected SR models is derived by applying the chain rule

 \begin{align}
  f_{\theta}(x) &= (I-A^+A)g_{\theta}(x) + A^+x\nonumber\\
  \\
  \frac{\partial f_{\theta}(x)}{\partial \theta} &=  \frac{\partial f_{\theta}(x)}{\partial g_{\theta}(x)}\frac{\partial g_{\theta}(x)}{\partial\theta} = (I-A^+A)\frac{\partial g_{\theta}(x)}{\partial\theta}
 \end{align}

Which is essentially the high-pass filtered version of the gradient of $g_{\theta}(x)$.

\section{Instance Noise\label{appdx:instancenoise}}

GANs are notoriously unstable to train, and several papers exist that try to improve their convergence properties \citep{salimans2016improved,radford2015unsupervised} via various tricks. Consider the following idealised GAN algorithm, each iteration consisting of the following steps:
\begin{enumerate}
\item we train the discriminator $D$ via logistic regression between $q_\theta$ vs $p_Y$, until convergence
\item we extract from $D$ an estimate of the logarithmic likelihood ratio $s(y) = \log \frac{q_\theta(y)}{p(y)}$
\item we update $\theta$ by taking a stochastic gradient step with objective function $\mathbb{E}_{y\sim q_\theta}s(y)$
\end{enumerate}
If $q_\theta$ and $p_Y$ are well-conditioned distributions in a low-dimensional space, this algorithm performs gradient descent on an approximation to the KL divergence, so it should converge. So why is it highly unstable in practical situations?

Crucially, the convergence of this algorithm relies on a few assumptions that don't always hold: (1) that the log-likelihood-ratio $\log \frac{q_\theta(y)}{p(y)}$ is finite, (2) that the Jensen-Shannon divergence $\operatorname{JS}[q_\theta\|p]$ is a well-behaved function of $\theta$ and (3) that the Bayes-optimal solution to the logistic regression problem is unique. We stipulate that in real-world situations neither of these holds, mainly because $q_\theta$ and $p_Y$ are concentrated distributions whose support may not overlap. In image modelling, distribution of natural images $p_Y$ is often assumed to be concentrated on or around a lower-dimensional manifold. Similarly, $q_\theta$ is often degenerate by construction. The odds that the two distributions share support in high-dimensional space, especially early in training, are very small. If $q_\theta$ and $p_Y$ have non-overlapping support (1) the log-likelihood-ratio and therefore KL divergence is infinite (2) the Jensen-Shannon divergence is saturated so its maximum value and is locally constant in $\theta$ and (3) there may be a large set of near-optimal discriminators whose logistic regression loss is very close to the Bayes optimum, but each of these possibly provides very different gradients to the generator. Thus, training the discriminator $D$ might find a different near-optimal solution each time depending on initialisation, even for a fixed $q_\theta$ and $p_Y$.

The main ways to avoid these pathologies involve making the discriminator's job harder. For example, in most GAN implementations the discriminator is only partially updated in each iteration, rather than trained until convergence. Another way to cripple the discriminator is adding label noise, or equivalently, \emph{one-sided label smoothing} as introduced by \citet{salimans2016improved}. In this technique the labels in the discriminator's training data are randomly flipped. However we do not believe these techniques adequately address all of the concerns described above.

\begin{figure}
  \center
  \includegraphics[width=0.7\linewidth]{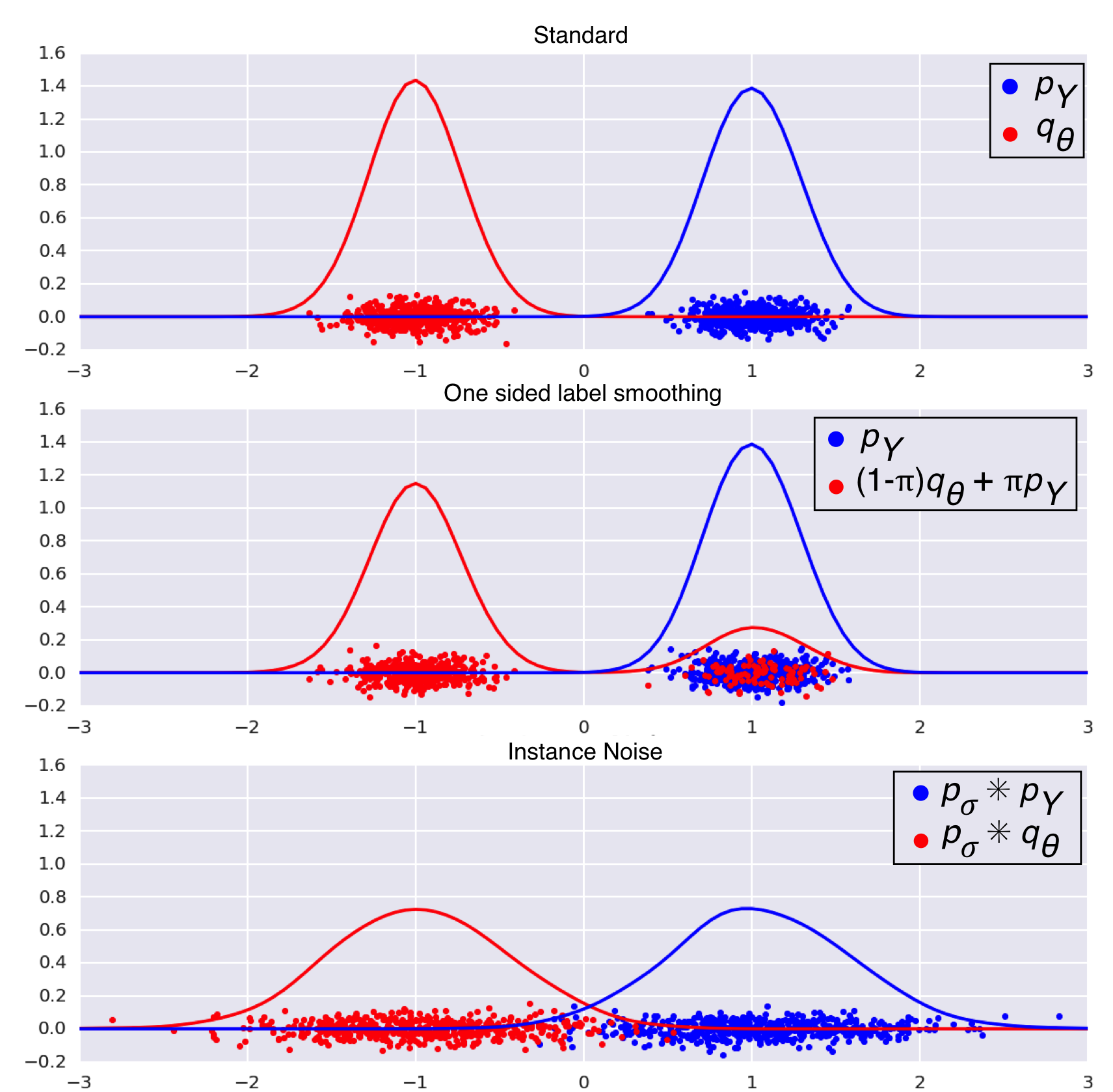}
  \caption{Illustration of samples from two non-overlapping perfectly distributions distributions in a,) with one-sided label smoothing in b) and instance noise in c). \emph{One side-label} smoothing shifts the optimal decision boundary but $p_Y$ still covers areas with no support in $q_{\theta}$. \emph{Instance Noise} broadens the support of both distributions without biasing the optimal discriminator.}
  \label{fig:instancenoise}
\end{figure}

In Figure~\ref{fig:instancenoise}a we illustrate two almost perfectly separable distributions. Notice how the large gap between the distributions means that there are large number of possible classifiers that tell the two distributions apart and achieve similar logistic loss. The Bayes-optimal classifier may not be unique, and the set of near-optimal classifiers is very large and diverse. In Figure~\ref{fig:instancenoise}b we show the effect of \emph{one sided label smoothing} or equivalently, adding label noise. In this technique, the labels of some real data samples $y\sim p_Y$ are flipped so the discriminator is trained thinking they were samples from $q_\theta$. The discriminator indeed has a harder task now, but all classifiers are penalised almost equally. As a result, there is still a large set of discriminators which achieve near-optimal loss, it's just that the near-optimal loss is now larger. Label smoothing does not help if the Bayes-optimal classifier is not unique.

Instead we propose to add noise to the samples, rather than labels, which we denote \emph{instance noise}. Using instance noise the support of the two distributions is broadened and they are no longer perfectly separable  as illustrated in Figure~\ref{fig:instancenoise}c. Adding noise, the Bayes-optimal discriminator becomes unique, the discriminator is less prone to overfitting because it has a wider training distribution, and the log-likelihood-ratio becomes better behaved. The Jensen-Shannon divergence between the noisy distributions is now a non-constant function of $\theta$.
Using instance noise, is easy to construct an algorithm that minimises the following divergence:
\begin{equation}
  d_\sigma(q_{\theta},p_Y) = \operatorname{KL}\left[p_{\sigma} \ast q_\theta \middle\| p_{\sigma} \ast p_Y\right],
\end{equation}
where $\sigma$ is the parameter of the noise distribution. Logistic regression on the noisy samples provides an estimate of $s_\sigma(x) = \log\frac{p_{\sigma} \ast q_\theta}{p_{\sigma} \ast p_Y}$. When updating the generator we have to minimise the mean of $s(\sigma)$ on noisy samples from $q_\theta$. We know that, if $p_\sigma$ is Gaussian, $d_\sigma$ is a Bregman-divergence, and that it is $0$ if and only if the two distributions are equal. Because of the added noise, $d_\sigma$ is less sensitive to local features of the distribution. We found that in our experiments instance noise helped the convergence of \affgan{}. We have not tested the instance noise in the generative modelling application. Because we don't have to worry about over-training the discriminator, we can train it until convergence, or take more gradient steps between subsequent updates to the generator. One critical hyper-parameter of this method is the noise distribution. We used additive Gaussian noise, whose variance we annealed during training. We propose a heuristic annealing schedule where the noise is adapted so as to keep the optimal discriminator's loss constant during training. It is possible that other noise distributions such as heavy-tailed or spike-and-slab would work better but we have not investigated these options.

\section{experimental details\label{appdx:experimental_details}}
\subsection*{Loss functions}

For the GAN models the generative and discriminative parameters were updated using Eqn.\ \eqref{eq:GANupdaterule}. For the models enforcing Eqn.\ \eqref{eq:io_consistency} using a soft-constraint we added an extra MAE loss term to the generative parameters $\ell_{MAE} = \frac{1}{N}\sum_i \| x_i - A\hat{y}_i \|$, where $i$ i runs over the number of data samples $N$.

The denoiser guided models were trained in a two step procedure. Initially we pre-trained a DAE to denoise samples from the data distribution by minimising

 \begin{align}
  \ell_{DAE} &= \frac{1}{N}\sum \|  y, f^{\sigma}_{DAE}(\tilde{y}) \|^2 \\
    \tilde{y} &= y + \epsilon,\ \ \epsilon\sim\mathcal{N}(0,\sigma I)\nonumber
 \end{align}

During training we anneal the noise level $\sigma$ and continuously save the model parameters of the DAE $f^{\sigma}_{(DAE)}$ trained at increasingly smaller noise levels. We then learn the parameters of the generator by following the gradient in Eqn.\ \eqref{eq:logPgrad} using the DAE to estimate $\frac{\partial}{\partial y} \log p(y)$

\begin{align}
 \frac{\partial}{\partial \theta}\mathbb{E}_{x}[\log p(\hat{y})] &= \mathbb{E}_{x} \left[ \frac{\partial}{\partial y} \log p(y) \cdot \frac{\partial}{\partial \theta} \hat{y} \right]\\
      &=  \mathbb{E}_{x} \left[ \frac{f^{\sigma}_{DAE}(\tilde{y}) - y}{\sigma^2} \cdot \frac{\partial}{\partial \theta} \hat{y} \right]\\
\theta_{i+1} &\leftarrow \theta_{i} + \alpha \frac{\partial}{\partial \theta}\mathbb{E}_{x}[\log p(\hat{y})]
\end{align}

Where $\alpha$ is the learning rate. During training we continuously load parameters of the DAE trained at increasingly low noise levels to get gradients pointing in the approximate correct direction in beginning of training while covering a large data space and precise gradients close to the data manifold in the end of the training.

For the density guided models we first pre-train a density model by maximising the tractable log-likelihood

 \begin{align}
  \mathcal{L}(y) &= \sum_j \log p(y_j|y_{<j})
 \end{align}

Where the joint density have been decomposed using the chain rule and $j$ runs over the pixels. Similar to the DAE we continuously save the parameters of the density model during training. We then learn the parameters of the generator by directly minimising the negative log-likelihood of the generated samples under the learned density model.

 \begin{align}
  \ell = -\mathcal{L}(\hat{y}) = -\mathcal{L}(f_{\theta}(x))
 \end{align}

\subsection*{2D swiss-roll}
The 2D target data $y = [y_1,y_2]$ was sampled from the 2D Swiss-Roll defined as:
\begin{align}
  \nu_1 &\sim \mathcal{N}(\mu_1,\sigma_1),\ \ \ \ \nu_2 \sim \mathcal{N}(\mu_2,\sigma_2)\\
  r &= 0.4\nu_1 + \nu_2\\
  y &= [\cos(\nu_1)*r, \sin(\nu_1)*r]
\end{align}
Where $\mu_1=10$, $\sigma_1=3$, $\mu_2=0$ and $\sigma_2=0.2$. The LR input was defined as $x=\frac{y_1+y_2}{2}$. The cross-entropy $\mathbb{H}[q_\theta,p_Y]$ were calculated by estimating the probability density function using a Gaussian kernel density estimator fitted to $50.000$ samples from a noiseless Swiss Roll density i.e. $\sigma_2=0$, and setting the bandwidth of each kernel to $\sigma_2=0.2$. All generator and discriminators were 2-layered fully connected NNs with 64 units in each layer. For the \affdenoiser{} model the DAE was a two layered NN with 256 units in each layer trained while annealing the standard deviation of the Gaussian noise from $0.5$ to $0.25$.

\subsection*{Image data}
For all image experiments we set $A$ to a convolution using a Gaussian smoothing kernel of size $9\times9$ using a stride of $4$ corresponding to $4\times$ down-sampling. $A^+$ were set to a convolution operation with $4^2$ kernels of size $5\times5$ followed by a reordering of the pixel with the output corresponding to $4\times$ up-sampling convolution as described in \citep{shi2016real}. The parameters of the $A^+$ was optimised numerically as described in Appendix \ref{sec:app_affproj}. All down-sampling were done using the $A$ projection. For all image models we used convolutional models using ReLU nonlinearities and batch normalization in all layers except the output. All generators used skip connections similar to \citep{huang2016densely} and a final sigmoid non-linearity was applied to output of the model which were either used directly or feed through the affine transformation layers parameterised by $A$ and $A^+$. The discriminators were standard convolutional networks followed by a final sigmoid layer.

For the grass texture experiments we used randomly extracted patches of data from high resolution grass texture images. The generators used 6 layers of convolutions with 32, 32, 64, 64, 128 and filter maps and skip connections after every second layer. The discriminators had four layers of strided convolutions with 32, 64, 128 and 256 filter maps. For the \affdenoiser{} model the DAE was a four layer convolutional network with 128 filter maps in each layer trained while annealing the standard deviation of the Gaussian noise from $0.5$ to $0.01$. The density model was implemented as a pixelCNN similar to \cite{van2016pixel} with four layers of convolution with 64 filter map with kernel sizes of 5, except for the first layers which used 7. The original PixelCNN uses a non-differentiable categorical distribution as the likelihood model why it can not be used for gradient based optimization. Instead we used a MCGSM as the likelihood model \citep{theis2015generative}, which have been shown to be a good density model for images \citep{Theis2012a}, using 32 mixture components and 32 quadratic features to approximate the covariance matrices.

For the CelebA experiments the datasets were split into train, validation and test set using the standard splitting. All images were center cropped and resized to $64\times64$ before down-sampling to $16\times16$ using $A$. All generators were 12 layer convolution networks with four layers of 128, 256 and 512 filter maps and skip connections between every fourth layer. The discriminators were  8 layer convolution nets with two layers of 128, 256, 512 and 1024 filter maps using a stride of 2 for every second layer.

 For the ImageNET experiments the 2012 dataset were randomly split into train, validation and test set with $10^4$ samples in the test and validation sets. All images below 20kB were then discarded to remove images with to low resolution.  The images were center cropped and resized to $128\times128$ before down-sampling to $32\times32$ using $A$. The generator were a 8 layer convolutional network with 4 layers of 128 and 256 filter maps and skip connections between every second layer. The discriminators were  8 layer convolution nets with two layers of 128, 256, 512 and 1024 filter maps using a stride of 2 for every second layer. To stabilise training we used Gaussian instance noise linearly annealed from an initial standard deviation of $0.1$ to $0$. We were unable to stable train models without this extra regularization.

\section{Additional results for Denoiser and Density guided Super-resolution\label{appdx:DG_models}}

\begin{figure}
  \center
  \includegraphics[width=0.8\linewidth]{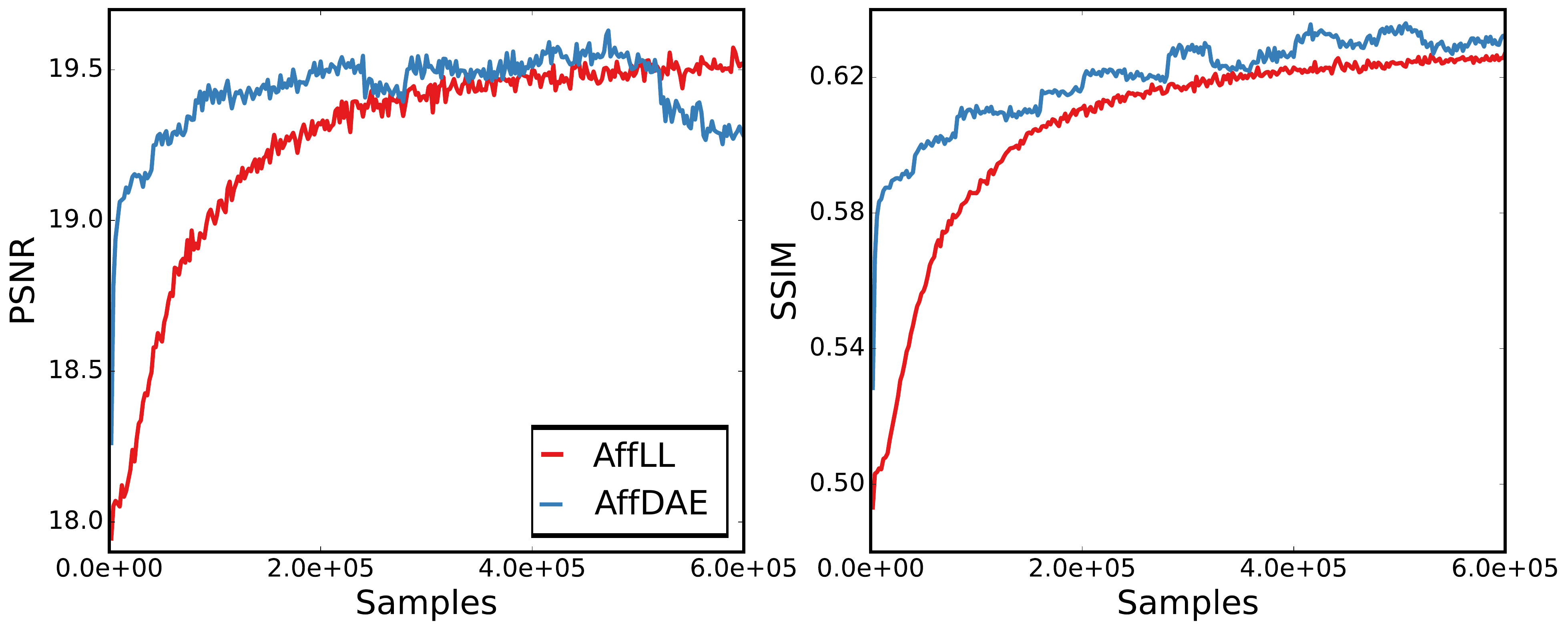}
  \caption{PSNR and SSIM results for the \affdenoiser{} and \affdensity{} models. Note that the step-like behaviour of the \affdenoiser{} model is due to change of the DAE model with continuously lower noise levels}
  \label{fig:density_denoise}
\end{figure}

Figure~\ref{fig:density_denoise} show the PSNR and SSIM scores during training for the \affdenoiser{} and \affdensity{} models trained on the grass textures.  Note that the models are converging, but as seen in Figure~\ref{fig:grass} the images are very blurry. For both models we had problems with diverging training. For the DAE models with high noise levels the gradients are only approximately correct but covers a large space around the data manifold whereas for small noise levels the gradients are more accurate in a small space around the data manifold. For the density model we believe a similar phenomenon is making the training diverge since for accurate density models the estimated density is likely very peaked around the data manifold making learning in the beginning of training difficult. To resolve these issue we started training using models with high noise levels or low log-likelihood values and then loaded model parameters during training with continuously smaller noise levels or better log-likelihood values. The effect of this can be clearly seen during training as the step like behavior of the \affdenoiser{} in Figure~\ref{fig:density_denoise}. We note that the density model used for training the \affdensity{} achieved a log-likelihood of $-4.10$ bits per dimension which is comparable to values obtained in \cite{theis2015generative} on a texture dataset. Further the \affdensity{} model achieved high log-likelihood values $>-3.5$ under this model suggesting that the density model is simply not providing an accurate enough representation of $p_Y$ to provide precises scores for training the \affdensity{} model.

\begin{figure}
  \center
  \includegraphics[width=1.0\linewidth]{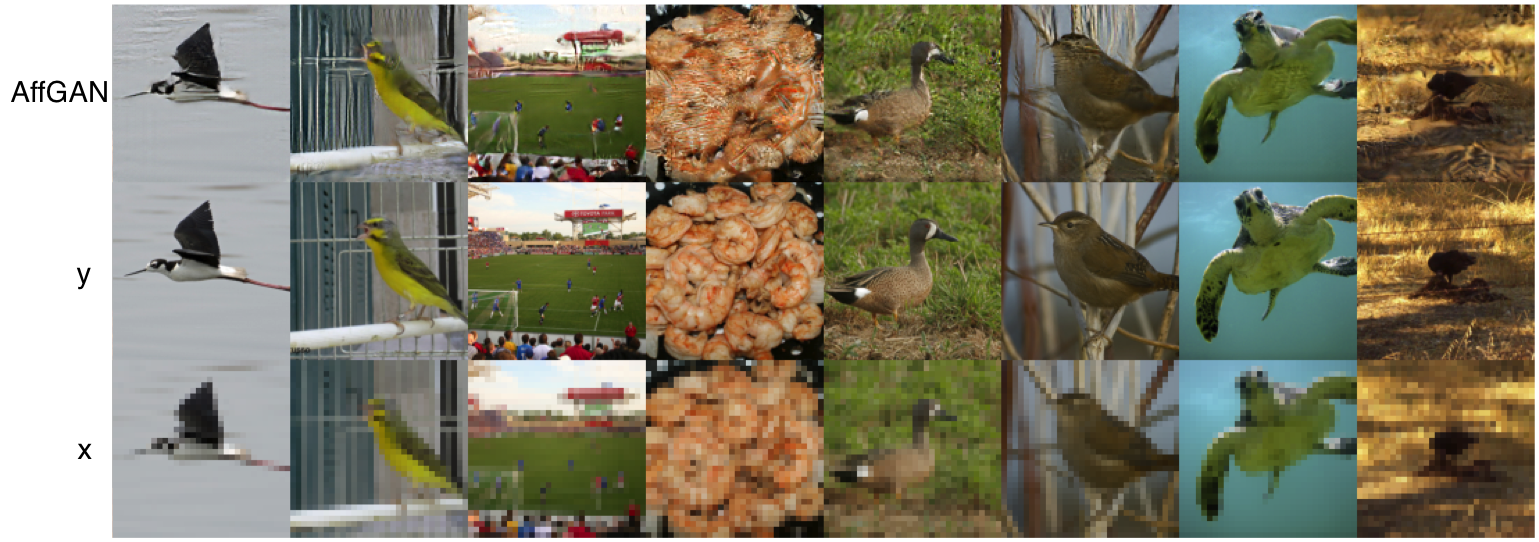}
  \caption{$4\times$ SR from $32\times32$ to $128\times128$ using \affgan{} on the ImageNET. \affgan{} outputs (top row), true HR images $y$ (middle row), model input $x$ (bottom row).}
  \label{fig:imagenet_panel2}
\end{figure}

\FloatBarrier
\section{Amortised variational inference using \affgan{}\label{sec:appendix_variational_affgan}}

Here we'll show that a stochastic extension of the \affgan{} model approximately minimises an amortised variational inference criterion as in e.\,g.\ variational autoencoders, which for the first time establishes a connection between adversarial methods of inferences and and variational inference. We introduce a variant of \affgan{} where, in addition to the LR data $x$, the \emph{generator} function also takes as input some independent noise variables $z$:
we establish a connection between GANs and amortised variational 
\begin{align}
z &\sim p_Z\\
\hat{y} &= \Pi^{A}_x f_\theta(x,z)
\end{align}
Similarly to how we defined $q_\theta$ in Section \ref{sec:theory_likelihood} we introduce the following notation:
\begin{align}
  q_{Y;\theta} &:= \mathbb{E}_{x\sim p_X} \mathbb{E}_{z\sim p_Z} \delta \left(y - \Pi^{A}_x f_\theta(x,z)\right)\\
  q_{Y\vert X;\theta} &:= \mathbb{E}_{z\sim p_Z} \delta\left(y - \Pi^{A}_x f_\theta(x,z)\right)\\
  q_{X,Y;\theta} &:= p_X \cdot q_{Y\vert X;\theta}
\end{align}

Here the affine projection ensures that under $q_{X,Y;\theta}$, $x$ and $y$ are always consistent. Therefore, under $q_{X,Y;\theta}$, the conditional of $x$ given $y$ is the same as the likelihood $p_{X\vert Y} = \delta(x - Ay)$ by construction and the following equality holds:
\begin{equation}
  q_{X,Y; \theta} = q_{Y;\theta} \cdot p_{X\vert Y} = p_{X} \cdot  q_{Y\vert X;\theta}
\end{equation}
Applying Bayes' rule to $p_{X\vert Y} = \frac{p_{Y\vert X}p_X}{p_Y}$ and substituting into the above equality we get:
\begin{equation}
  q_{Y;\theta} \cdot p_{Y\vert X=Ay} = p_{Y; \theta} \cdot q_{Y\vert X=Ay;\theta}
\end{equation}
The $\operatorname{KL}$ divergence that the \affgan{} objective minimises can now be rewritten as.
\begin{align}
  \operatorname{KL}\left[q_{Y;\theta}\middle\|p_Y\right] &= \mathbb{E}_{q_{Y;\theta}} \log \frac{q_{Y;\theta}(y)}{p_Y(y)}\\
  &= \mathbb{E}_{q_{X,Y;\theta}} \log \frac{q_{Y;\theta}(y)}{p_Y(y)}\\
  &= \mathbb{E}_{q_{X,Y;\theta}} \log \frac{q_{Y\vert X;\theta}(y\vert x)}{p_{Y\vert X}(y\vert x)}\\
  &= \mathbb{E}_{p_X} \operatorname{KL}\left[q_{Y\vert X;\theta}\middle\|p_{Y\vert X}\right]
\end{align}

Therefore we can conclude that the \affgan{} algorithm described in Section\ \ref{sec:GAN} approximately minimizes the following amortised variational inference criterion:
\begin{equation}
  \argmin_\theta \operatorname{KL}\left[q_{Y;\theta}\middle\|p_Y\right] = \argmin_\theta \mathbb{E}_{x\sim p_X} \operatorname{KL}\left[q_{Y\vert X;\theta}\middle\|p_{Y\vert X}\right],
\end{equation}
and in doing so it only requires samples from $P_Y$ and $P_X$.

\end{document}